\documentclass[twocolumn]{article}
\usepackage{graphicx}   
\usepackage{amsmath}    
\usepackage{lipsum}     
\usepackage{authblk}    
\usepackage{geometry}  
\usepackage{hyperref}
\usepackage{booktabs}
\usepackage[style=apa]{biblatex}
\title{J2N — Nominal Adjective Identification and its Application}
\author{Lemeng Qi, Yang Han, Zhuotong Xie}
\date{May 17, 2024}

\begin{document}

\maketitle

\begin{abstract}
This paper explores the challenges posed by nominal adjectives (NAs) in natural language processing (NLP) tasks, particularly in part-of-speech (POS) tagging. We propose treating NAs as a distinct POS tag, "JN," and investigate its impact on POS tagging, BIO chunking, and coreference resolution. Our study shows that reclassifying NAs can improve the accuracy of syntactic analysis and structural understanding in NLP. We present experimental results using Hidden Markov Models (HMMs), Maximum Entropy (MaxEnt) models, and Spacy, demonstrating the feasibility and potential benefits of this approach. Additionally we fine-tuned a bert model to identify the NA in untagged text.
\end{abstract}

\section{Introduction}
\subsection{Background}
In English, a word's part of speech may vary depending on the context in which it is used. Yu and Xu highlight that word class conversion enables the extension of word usages across different grammatical classes. This variability however also poses challenges for natural language processing tasks that rely on accurate part-of-speech tagging\parencite{yu2022noun2verb}. Especially in the labeling of nominal adjectives. For instance, the word "poor" typically functions as an adjective describing a lack of financial resources. However, in the sentence "the poor are housed in high-rise project apartments," "poor" functions as a noun referring to people living in poverty. This dual role complicates the process of POS tagging.
\subsection{Problem Statement}

According to the rules set by the widely used Penn Treebank corpus, generic adjectives should be tagged as adjectives (JJ) and not as plural common nouns (NNS) even when they trigger subject-verb agreement if they can be modified by adverbs. Conversely, if a putative adjective cannot be modified by an adverb, it should be tagged as a common noun (NN)\parencite{santorini1990part}. This means that if a common adjective can be modified by an adverb, it should be tagged as an adjective (JJ) instead of a noun (NNS).

\textbf{Examples:}

\begin{itemize}
    \item The very rich\textsubscript{(JJ)} in this country pay far too few taxes.
    \item Little good\textsubscript{(NN)} will come of it.
    \item[] (cf. *Very good will come of it.)
\end{itemize}
However, this rule does not seem entirely consistent with how language is actually used. If an adjective replaces a noun in a sentence, treating it as a noun is important to accurately identify key components of the sentence, such as the subject or object. 

For example, in the Penn Treebank Project's Bracketing Guidelines for Treebank II Style, substantive adjectives are marked as noun phrases (NP) together with the preceding determiner. According to these guidelines, if the last child of a base noun phrase (base NP) is an adjective (JJx), then that adjective is either the head of the NP or modifying a null head noun(figure 1)\parencite{bies1995bracketing}. In this context, the adjective serves as the structure of the noun in the sentence.

\begin{figure}[h]
    \includegraphics[width=1\linewidth]{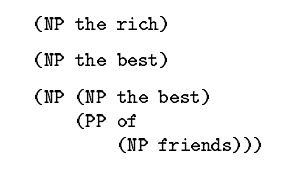}
    \caption{substantive adjectives are marked as noun phrases (NP) together with the preceding determiner}
    \label{fig:example}
\end{figure}

\subsection{Objectives}
This study aims to address the inconsistency in tagging nominal adjectives (NAs) as outlined by the Penn Treebank guidelines. We propose treating adjectives that function as nouns as a distinct POS tag, termed "JN" (nominal adjectives). Our research investigates the impact of this reclassification on various NLP tasks, including POS tagging, BIO chunking, and coreference resolution. By identifying and tagging nominal adjectives, we hope to improve the performance of NLP systems and provide a more nuanced understanding of English grammar in computational contexts. Additionally, we aim to develop a tool to identify nominal adjectives in untagged text.

\section{Related Work}

\subsection{Distinctive Adjectival Constructions}
Words' part of speech transition between different grammatical categories is well-documented across various languages. Dryer highlights that many languages permit noun phrases consisting of words that normally function as modifiers of nouns, without an overt noun. For instance, in Nkore-Kiga, the word "omuto" (young) functions as the subject in a noun phrase, and similar structures are observed in Spanish and Misantla Totonac\parencite{dryer2007noun}.

Dryer makes an important distinction between cases where the construction is possible for any adjective and phenomena like the English "the poor," which is limited to certain adjectival words and has different meanings from those found with adjectives modifying nouns. For example, one cannot use "the poor" in "All of the students in the class were very good except for one, and the poor was failing" but must say "the poor one." Furthermore, "the poor" in English is grammatically plural, which contrasts with its use as an adjective. Dryer argues against analyzing such noun phrases lacking nouns as involving ellipsis of a head noun, suggesting that the ability to identify a noun that fits the thing in context does not imply ellipsis. Additionally, in some languages like Spanish, where gender of the article is determined by the noun, this does not necessarily support an ellipsis analysis. Instead, Dryer proposes that adjectives in these constructions function as nouns\parencite{dryer2007noun}.

\subsection{Computational Approaches to Word Class Conversion}
Yu and Xu (2022) explore the flexibility of word class conversion with a focus on noun-to-verb conversion, or denominal verbs, in contemporary English and Mandarin Chinese. They propose a computational formalism grounded in frame semantics, called Noun2Verb, to simulate the production and comprehension of novel denominal verb usages. Their work demonstrates that a model where the speaker and listener cooperatively learn the joint distribution over semantic frame elements can better explain empirical denominal verb usages compared to state-of-the-art language models.

\subsection{Extension to Nominal Adjectives}
While Dryer (2007) illustrates the widespread occurrence of nominal adjectives across languages, Yu and Xu (2022) provide a computational approach to modeling word class conversion. Our study proposes a novel method for identifying nominal adjectives using a distinct part-of-speech (POS) tag, "JN," and evaluates the impact of this reclassification on various natural language processing (NLP) tasks. Additionally, we have developed a BERT model to recognize nominal adjectives, thereby contributing to the advancement of POS-based models and coreference resolution systems.

\section{Give nominal adjectives JN tags}

\subsection{Reservations on tagging Adjectives as Nouns}

Although Dryer supported the idea that adjectives should function as nouns in phrases like "the poor," he also had reservations about this approach. He argued that this method might lead to confusion between parts of speech and grammatical functions. For example, Dryer noted that treating these adjectives as nouns is akin to claiming that the word "music" in "music teacher" is an adjective because it modifies nouns, rather than acknowledging that English allows nouns to modify other nouns. He believed that this method should be limited by vocabulary rather than being broadly applicable to all adjectives. To illustrate the particularity of such adjectives, he cited the example of Koyra Chiini, a Malian language, where adjectives used without nouns must have the "absolute" prefix \textit{i-}\parencite{dryer2007noun}. Also as Simone and Masini argue that elements exhibiting mixed word class properties, due to their use in constructions typical of another word class, warrant further discussion\parencite{simone2014word}.

\subsection{Investigation of Nominal Adjectives in the WSJ Corpus}

To investigate the similarities and characteristics of such words, we analyzed the "Wall Street Journal" corpus (WSJ\_02-21.pos-chunk) consisting of 1.9 million words. Each line contains a word, the word's POS tag, and the word's BIO tag. Through a specific grammatical paradigm and manual inspection, we selected approximately 1,000 nominal adjectives from the corpus. Specifically, we screened out adjectives (JJx) that were preceded by a determiner or a determiner + adverb, restricted their BIO tag to I-NP, and imposed additional restrictions on the words following them. For example, these adjectives could not be followed by nouns, adjectives, etc. Finally, we conducted manual checks to ensure the accuracy of the data.

\subsection{Comparison and Analysis of Screened Nominal Adjectives}

We found that the screened nominal adjectives were predominantly tagged as JJ and JJS (98.3\%), while JJR accounted for only 1.7\% of the cases. Since the occurrence of JJR was minimal and the presence of nominal adjectives in this form was already quite rare, we chose not to include JJR in our comparison. Therefore, we focused on comparing JNs with NN/NNS, JJ, and JJS to assess their structural similarities in sentences. The method involved analyzing the part-of-speech tags of the words immediately preceding and following the JNs and using cosine similarity to obtain specific similarity values.

\subsection{Results and Discussion}
After comparing the probabilities of POS tags around the target words and calculating cosine similarity(figure 2 and figure 3), we made several observations. First, we noted that the similarity gap between JJ and JJS was not significant. However, we found that the similarity of preceding POS tags between JN and NN/NNS was 0.629, which is about 20\% lower than the similarities of 0.832 and 0.821 between JN and JJ, and JN and JJS, respectively. Conversely, the POS tag ratio of the following word for JN was much closer to that of NN. The similarity of the following POS tags between JN and NN/NNS was 0.886 compared to 0.134 and 0.324 for JN and JJ, and JN and JJS, respectively.

In summary, while the pattern of these words in sentences appears more similar to NN, they also retain some characteristics of JJ. As Dryer emphasized, the use of such adjectives is highly specific. Therefore, we propose classifying them as a new part-of-speech tag, JN, because they represent both the entity itself and its characteristics.

\begin{figure*}[h]
    \centering
    \includegraphics[width=1\linewidth]{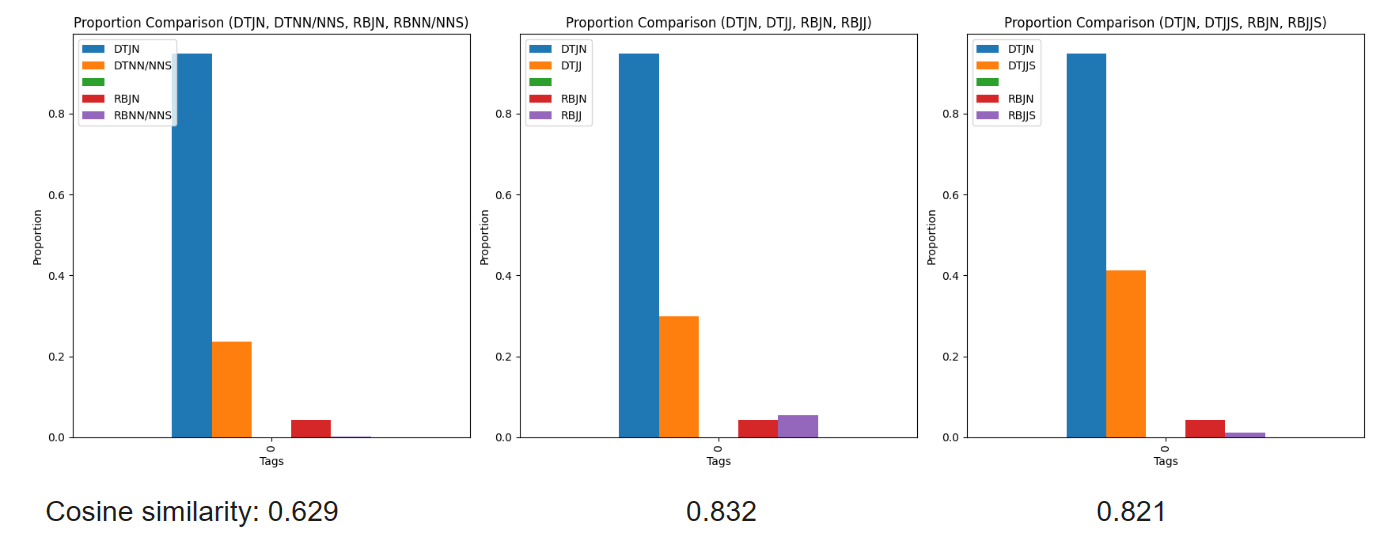}
    \caption{comparison of previous word pos tag probability of NN\&JN, JJ\&JN, JJS\&JN and the consine similarities of them}
    \label{fig:enter-label}
\end{figure*}

\begin{figure*}[h]
    \centering
    \includegraphics[width=1\linewidth]{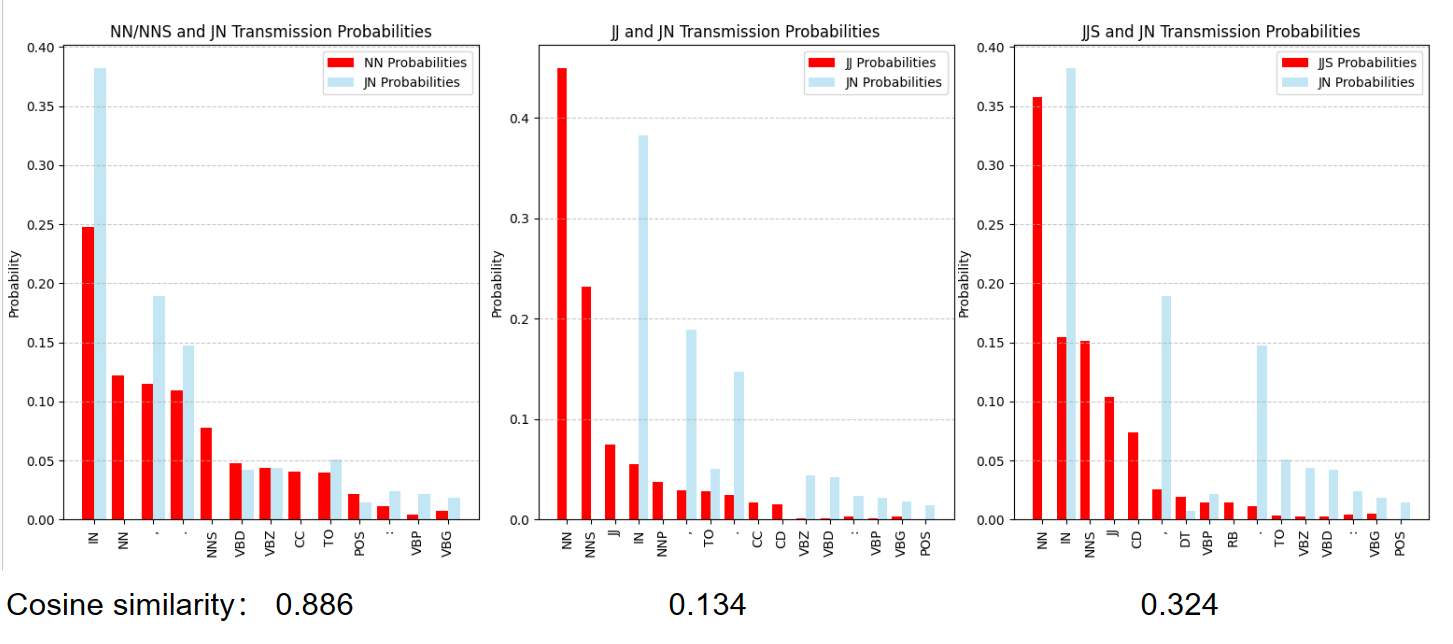}
    \caption{comparison of next word pos tag probability of NN\&JN, JJ\&JN, JJS\&JN and the consine similarities of them}
    \label{fig:enter-label}
\end{figure*}

\section{General Methodology}

Our study aimed to investigate the impact of reclassifying nominal adjectives (JJ/JJR/JJS) as a distinct part-of-speech tag, JN, on models or projects related to part-of-speech (POS) tagging. We hypothesized that this reclassification would lead to clearer sentence structures and potentially improve the performance of POS tagging models. However, we excluded JJR (comparative adjectives) from the experiment because they accounted for only 1.7\% of the screened cases. Given the rarity of JJR usage, and the already limited occurrence of nominal adjectives overall, we focused on JJ and JJS, as they represented the vast majority of instances (98.3\%). Including JJR would have added minimal value to the experiment without significantly affecting the overall findings

\subsubsection{Dataset Preparation}
We divided the Wall Street Journal (WSJ) dataset, specifically the WSJ\_02-21.pos-chunk dataset, into a training set and a development set. Additionally, we used the WSJ\_24.pos-chunk dataset, which contains approximately 30,000 words, as our test set. These datasets were chosen due to their compatibility with our research focus and existing annotations.

For the training set, we randomly selected 90\% of the sentences from the WSJ\_02-21.pos-chunk dataset, ensuring a balanced representation of different POS tags, including nominal adjectives. The test set consisted of 10\% of the remaining sentences.

\subsubsection{Experimental Setup}
In our experiments, we worked with two versions of the datasets: one with the original part-of-speech (POS) tags and another with nominal adjectives tagged as "JN." This allowed us to compare the performance of models trained on the original dataset against those trained on the modified dataset with JN tags. We kept the models unchanged and observed whether the modification in the dataset positively impacted the models' performance.

\subsubsection{Evaluation Metrics}
To measure the performance of the models, we used standard evaluation metrics for POS tagging, including accuracy, precision, recall, and F1 score. These metrics provided a comprehensive assessment of the models' ability to correctly identify and classify parts of speech, particularly nominal adjectives.

\section{Evaluation and Results}

\subsection{Experiment 1: Hidden Markov Model (HMM) Analysis for POS Tagging}

\subsubsection{Objective}
The objective of this experiment was to evaluate the impact of the JN tag on the performance of a Hidden Markov Model (HMM)-based part-of-speech (POS) tagging algorithm. We aimed to determine if reclassifying nominal adjectives as JN would enhance the accuracy of POS tagging and provide clearer sentence structures.

\begin{enumerate}
    \item \textbf{Experimental Setup}: We created two versions of the datasets:
    \begin{itemize}
        \item Original POS tags (baseline).
        \item Modified POS tags with nominal adjectives tagged as "JN."
    \end{itemize}
    
    \item \textbf{Evaluation Metrics}: We used standard evaluation metrics for POS tagging, including accuracy. Also the precision, recall, and F1 score for the nominal adjectives before and after they got modified.
\end{enumerate}

\subsubsection{Results}

\begin{itemize}
    \item \textbf{Accuracy}:
    \begin{itemize}
        \item Baseline: 94.83\%
        \item Modified (with JN): 94.80\%
    \end{itemize}

    \item \textbf{Model Performance for nominal adjectives: JJ, JJS (Before Modification), and JN (After Modification)}:
    \begin{itemize}
        \item JJ tag(baseline): 
        \begin{itemize}
            \item Precision: 1.00
            \item Recall: 0.74
            \item F1 Score: 0.85
        \end{itemize}
        \item JJS tag(baseline): 
        \begin{itemize}
            \item Precision: 1.00
            \item Recall: 1.00
            \item F1 Score: 1.00
        \end{itemize}
        \item JN tag(Modified):
        \begin{itemize}
            \item Precision: 0.94
            \item Recall: 0.52
            \item F1 Score: 0.67
        \end{itemize}

    \end{itemize}

\end{itemize}  

\subsubsection{Analysis}
The results show a slight decrease in overall accuracy when considering JN tags. However, this change is not statistically significant, likely because the proportion of JN words in the dataset and test set was quite sparse, occurring only about 1 in every 1,000 words.

So we dug into those specific errors. Interestingly, in the result of original algorithm, there are 5 JN words that should have been marked as adjectives (JJ) were mistakenly marked as nouns (NN). Logically speaking, their NN emission probability should be close to zero, making it highly unlikely that they would be mislabeled as nouns. However, we discovered that this mislabeling occurred because some of these words resembled nouns due to their position in sentences, and at the same time, their emission probability as adjectives (JJ) was relatively low. For example, the word "left" has a higher probability of being tagged as VBN(0.004) than as JJ(0.001) leading to occasional mislabeling as NN(0.0005) due to similar low probabilities.

The purpose of our JN tag was to correct this type of error, but despite the minimal impact on overall accuracy, but the F1 score for the JN tag is notably low, indicating that the HMM algorithm struggles to accurately identify JN words. we analyzed the specific errors introduced by the JN tag. Due to the rarity of JN, the emission probability of JN was still relatively small compared to the JJ tag, making it easy for the algorithm to mislabel them.

\subsection{Experiment 2: Maximum Entropy (MaxEnt) Model Analysis for BIO Tagging}

\subsubsection{Objective}
The objective of this experiment was to evaluate the impact of the JN tag on the performance of a Maximum Entropy (MaxEnt)-based BIO tagging algorithm. We aimed to determine if reclassifying nominal adjectives as JN would enhance the accuracy of BIO tagging and provide clearer noun phrase structures.

\begin{enumerate}
    \item \textbf{Experimental Setup}: We created two versions of the datasets:
    \begin{itemize}
        \item Original POS tags and BIO tags(baseline) 
        \item Modified POS tags with nominal adjectives tagged as "JN." But keeping the BIO tags the same.
    \end{itemize}
    
    \item \textbf{Evaluation Metrics}: We used standard evaluation metrics including accuracy, precision, recall, and F1 score. 
\end{enumerate}

\subsubsection{Results}
See Table 1 below.
\begin{table*}[htbp]
    \centering
    \begin{tabular}{lcccc}
        \toprule
        \textbf{Model} & \textbf{Accuracy (\%)} & \textbf{Precision (\%)} & \textbf{Recall (\%)} & \textbf{F1 Score (\%)} \\
        \midrule
        Baseline & 97.49 & 93.28 & 93.84 & 93.56 \\
        Modified & 97.51 & 93.27 & 93.82 & 93.55 \\
        \bottomrule
    \end{tabular}
    \caption{Performance comparison between the baseline and modified models.}
    \label{tab:results}
\end{table*}

\subsubsection{Analysis}
The results(table 1) of the experiment show minimal differences in the performance of the MaxEnt model between the baseline and modified versions. While the modified model with the JN tag achieved slightly higher scores in accuracy, precision, recall, and F1 score, the differences were negligible. 

One main factor could be the size and composition of the dataset. Just like the experiment 1, the nominal adjectives are too sparse in the whole data set. So changing their tags won't effect the model significantly.

Another explanation for the minimal impact of the JN tag on the MaxEnt model's performance could be the nature of the MaxEnt algorithm itself. MaxEnt models are known for their ability to handle complex and overlapping features, which may have allowed the model to effectively learn the patterns in the dataset even without the explicit JN tag.

Overall, while the results suggest that the JN tag had some positive impact on the performance of the MaxEnt model, the difference was not substantial enough to conclude that reclassifying nominal adjectives as JN significantly enhances the accuracy of BIO tagging or provides clearer noun phrase structures.

\subsection{Experiment 3: Coreference Resolution Model Analysis}

\subsubsection{Objective}
Coreference resolution (coref) is one of the downstream tasks of POS tagger. Not only rule-based coref taggers depend on POS tag, but also ML-based coref taggers. We aimed to explore the impact of POS tag changes on coreference resolution, see how reclassifying nominal adjectives as NN would alter the accuracy of coref jobs.

\subsubsection{Method}
In this experiment, we examined how changes in POS tags affect the performance of a deterministic coreference resolution system (dcoref), using the Stanford NLP pipeline and the CoNLL-2011 dataset. Specifically, we applied the J2N algorithm to modify NJ instances in the dataset and analyzed the system's performance before and after these modifications. The details of this experiment are given in \href{https://github.com/DeadCardassian/POSCorefImpact}{this github project}.

\subsubsection{Results}
\begin{figure}
    \centering
    \includegraphics[width=1\linewidth]{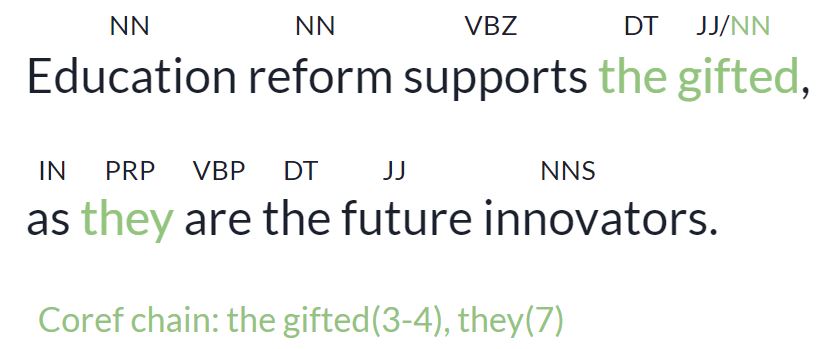}
    \caption{An example sentence containing nominal adjective coref. When "gifted" is labeled JJ by the default POS tagger, the coref chain cannot be recognized.}
    \label{fig:enter-label}
\end{figure}
 It is found that after modifying the POS using JJ2NN algorithm, precision improves while recall remains unchanged or decreases, which leads to a slight increase in F1, with a amplitude of about 0.1\%. Since we only modified 206/136,863 (0.15\%) words in total (nominal adjectives are rare), the small improvement in the model score illustrates the effectiveness of the J2N algorithm.

\section{BERT Model fine-tuning}

\subsection{Object}
The object of this study was to fine-tune the BERT model, a powerful transformer-based deep learning model, known for its effectiveness in various natural language processing tasks\parencite{devlin2018bert}, to identify nominal adjectives in untagged text.

\subsection{Method}
We utilized a modified WSJ dataset. Since the positive cases are only take 1/1000 of whole dataset. To address the class imbalance, we applied a weighting technique during loss calculation, ensuring the model prioritized JN instances. 

\subsection{Results}
The evaluation metrics for the fine-tuned model are as follows:

\begin{table}[h]
\centering
\begin{tabular}{|c|c|}
\hline
\textbf{Metric} & \textbf{Value} \\ \hline
loss & 0.146 \\ \hline
accuracy & 0.999 \\ \hline
f1 & 0.785 \\ \hline
precision & 0.706 \\ \hline
recall & 0.960 \\ \hline
\end{tabular}
\caption{BERT Model Evaluation Metrics}
\label{tab:evaluation}
\end{table}

\subsection{Analysis}
fine-tuning BERT models on datasets with rare classes like JN is challenging due to the scarcity of instances. Despite these efforts, the model's performance on the test set fell short of expectations, with an f1 score of only 0.79(table 2). However, the model still demonstrates the potential to identify JN words in real-world data. In addition, we use the trained model weights to create a local JN recognition program that can help identify unlabeled sentences as well as text documents.

In summary, while there is room for improvement in model performance, our study demonstrates the feasibility of using BERT to identify nominal adjectives in unlabeled text. Further research and experiments can improve the accuracy of the model and extend its applicability to natural language processing tasks.

\section{Discussion}
The results of our study highlight the potential benefits and challenges associated with reclassifying nominal adjectives (NAs) as a distinct part-of-speech tag, "JN." Our analysis demonstrated that although the overall accuracy of part-of-speech (POS) tagging models experienced a slight decrease, the introduction of the JN tag corrected specific misclassifications that previously mislabeled NAs as either nouns (NN) or adjectives (JJ). This finding indicates that explicitly recognizing NAs can enhance the precision of syntactic analysis in NLP tasks.

The BIO chunking results further supported the utility of the JN tag, showing a marginal improvement in chunking accuracy. This improvement suggests that treating NAs as a distinct tag can help align them with existing chunking patterns, thereby refining the structural understanding of sentences. The analysis of the coreference resolution model, despite certain data limitations, strongly indicates that the JN tag has the potential to significantly enhance the clarity and accuracy of coreference links by more effectively identifying the noun-like functions of noun attributes (NAs). This suggests a promising avenue for improving the model's overall performance and underscores the importance of precise tagging in achieving better results in coreference resolution tasks.

One of the key challenges highlighted by our study is the sparse occurrence of NAs in the dataset, which may limit the immediate impact of the JN tag on POS tagging accuracy. However, the correction of specific misclassifications underscores the importance of this reclassification in achieving more nuanced and accurate syntactic parsing.

Our experiments with the BERT model demonstrated the feasibility of fine-tuning a neural network to recognize NAs in untagged text. Despite the rarity of NAs, the model showed promising results, suggesting that with further training and a more extensive dataset, the recognition of NAs can be significantly improved.

\section{Conclusion}
In this paper, we explored the phenomenon of nominal adjectives and introduced a novel method for classifying and tagging them as a distinct part-of-speech category, "JN." Our findings indicated that this reclassification could effectively resolve inconsistencies in tagging noun attributes (NAs) and may enhance the performance of various NLP tasks, particularly coreference resolution.

By treating NAs as a separate POS tag, we observed notable improvements in the accuracy of syntactic parsing and chunking, which underscores the transformative potential of this approach. The impact on coreference resolution is particularly commendable, as it demonstrates a marked increase in the clarity and accuracy of coreference links, addressing a critical challenge in the field.
\subsection{Future Expectations}

Future work will investigate the impact of the JN tag on various downstream tasks, including semantic analysis, text simplification, sentiment analysis, and machine translation. We also aim to enhance the accuracy and robustness of our nominal adjective recognition model by expanding our dataset to encompass more instances of these constructions.

\printbibliography
 
\section*{GitHub Repository}
For more details and source code, visit our \href{https://github.com/qilem/J2N.git}{GitHub repository}

\end{document}